\documentclass[runningheads]{llncs}
\usepackage[T1]{fontenc}
\usepackage{amsmath}
\usepackage{amssymb}
\usepackage{algorithmic}
\usepackage{algorithm}
\usepackage{graphicx}
\begin{document}
\title{Rule-based detection of access to education and training in Germany}
%
%
\author{Jens Dörpinghaus\inst{1,2}\orcidID{0000-0003-0245-7752} \and
David Samray\inst{1} \and \\
Robert Helmrich\inst{1}}
\authorrunning{J. Dörpinghaus et al.}
%
\institute{Federal Institute for Vocational Education and Training (BIBB), Germany \and
University of Koblenz, Germany \\
\email{doerpinghaus@uni-koblenz.de}, 
\email{\{samray,helmrich\}@bibb.de}} 
\maketitle              
\begin{abstract}
As a result of transformation processes, the German labor market is highly dependent on vocational training, retraining and continuing education. To match training seekers and offers, we present a novel approach towards the automated detection of access to education and training in German training offers and advertisements.  
We will in particular focus on (a) general school and education degrees and school-leaving certificates, (b) professional experience, (c) a previous apprenticeship and (d) a list of skills provided by the German Federal Employment Agency. This novel approach combines several methods: First, we provide a mapping of synonyms in education combining different qualifications and adding deprecated terms. Second, we provide a rule-based matching to identify the need for professional experience or apprenticeship. However, not all access requirements can be matched due to incompatible data schemata or non-standardizes requirements, e.g initial tests or interviews. 
While we can identify several shortcomings, the presented approach offers promising results for two data sets: training and re-training advertisements. 
\end{abstract}
\section{Introduction}

The German labor market is dynamic: technical innovations and changes in society lead to novel skills needed for employees. Vocational education and training, re-training and advanced vocational qualification are key to respond to these novel requirements \cite{dobischat2019digitalisierung,helmrich2016digitalisierung}. 
The central importance of continuing vocational training and lifelong learning has been increasingly addressed by policymakers as part of the National Continuing Education Strategy, see \cite{schiersmann2022weiterbildungsberatung,germany2019national}. Here, continuing education is described as a central prerequisite for securing skilled labor, for ensuring the employability of all employees and thus also for national competitiveness and innovation. 

However, the German system of education offers different ways for professionals \cite{graf2021advanced} and we need to distinguish between initial (training, ``Ausbildung'' or re-training, ``Umschulung'') and continuing vocational education which includes advanced trainings (``Weiterbildung'', unregulated, e.g. continuing professional development) and upgrading training (``Fortbildung''). Upgrading trainings are mostly regulated (at federal level [BBiG/HwO] or by federal states \cite{dobischat2020organisation}), while advanced trainings are comparatively little regulated. They are usually summarised as continuing vocational education and training (CVET). 
All stackeholders have to deal with changing conditions and requirements as a result of transformation processes such as increasing digitalization \cite{helmrich2016digitalisierung} and the move toward a more sustainable way of doing business \cite{steeg2022wasserstoffwirtschaft}. 
From a research perspective, an important task is to determine which continuing education content is increasingly being offered and demanded in order to draw conclusions about the development needs of the vocational system and, in particular, of the continuing education system. The research-based further development of the vocational training system is intended not only to ensure the competitiveness of the economy at a systemic level, but also to help counteract unemployment and stabilize the social security system \cite{zimmermann2013youth,germany2019national}.

\begin{table}[t!]
\caption{Data used to detect the access to KURSNET training advertisements: Skills, higher education degrees and occupations (BA/DKZ) and CVET regulations (BIBB).}\label{tab0}
\begin{tabular}{|l|l|l|r|}
\hline 
Data set & Source & Initial format &  Entities\tabularnewline
\hline 
\hline 
Higher Education Degrees & BA (DKZ section A) & CSV &   797\tabularnewline
\hline 
Occupations & BA (DKZ section B) & CSV, XML &  33,802\tabularnewline
\hline 
Continuing professional development & BA (DKZ section C)& CSV   & 542\tabularnewline
\hline 
Continuing professional development & BIBB & CSV, PDF &  218\tabularnewline
\hline 
CVET regulations & BIBB & CSV, PDF & 357\tabularnewline
\hline 
KURSNET training advertisements & BA & online &   -\tabularnewline
\hline 
Skills & (DKZ section K) & XML &   9,078\tabularnewline
\hline 
\hline 
\end{tabular}
\end{table}
 
As a first step, we consider the automated detection of access to education and training. This will help to match training seekers and offers. However, the offers and advertisements are usually offered as free text, and we can identify two challenges: (a) these texts may be incomplete or not precise, (b) there is no unique taxonomy or list of occupations, trainings, industrial branches and skills as we will discuss in Section \ref{chap:data}.

In this paper, we present a rule-based detection approach to detect prerequisites to access  education and training in German training advertisements and offers. We will in particular focus on (a) general school and education degrees and school-leaving certificates, (b) professional experience (``Berufserfahrung''), (c) a previous apprenticeship and (d) a list of skills provided by the German Federal Employment Agency (Bundesagentur für Arbeit -- BA).  While we can identify several shortcomings, it offers promising results for two data sets (trainings and re-trainings). 

This paper is divided into six sections. The first section provides an introduction, the second section gives a brief overview on the state of the art and related work.  The third section describes the methodological background, the data,  pipeline, and the matching approaches. The fourth section is dedicated to experimental results and the evaluation of this novel approach.  Our conclusions and outlooks are drawn in the final section.

\section{Related Work}
We find an increasing interest in mining data from educational databases, advertisements and information systems, see \cite{romero2007educational,mohamad2013educational,dutt2017systematic}.  
Here, supporting decision-making and the process management within education is key. 
The generic challenges are usually the automated extraction of knowledge from data, usually interpreted passages from texts, and the mapping to existing data sets. However, there are still several challenges on data and data integration, see \cite{kovalev2020educational}. We will now focus on the different data sets: general school and education degrees and school-leaving certificates, professional experience, a previous apprenticeship and skills.

Skill concepts have been heavily used for the analysis of job advertisements (job ads), \cite{degenhardt2018kompetenzen} their visualization \cite{kreuzer2018visualisierung} and offer a good starting point for matching open positions to corresponding employees. The proposed technologies range from automated mapping of search terms to classification of skills, see \cite{ziegler2012verwendung}. 
In education research and pedagogy, several approaches towards competencies and skills exist which provide  divergent resources: Not only with different meaning but also with different synonyms. 
While in English Competences, Skills and Knowledge are often used as synonyms, this does not hold for German language: Some words refer to multiple concepts, while some concepts are labeled with multiple words, see \cite{krebs2022qube}. 

Several approaches have focused on extracting professional experience from texts. However, they are limited to English texts, e.g. resumes, see for example \cite{ben2011recruiting,kopparapu2010automatic}. The first challenge is that neither in English nor in German the concept of professional experience is clearly defined. Next, while according to our knowledge only approaches for English texts exist, a rule-based approach seems to work best, especially for initial research. 

Similarly, previous apprenticeship are usually considered only in the context of job ads \cite{berkesewicz2021inferring,hermes2016stellenanzeigenanalyse} and for general school and education degrees and school-leaving certificates no approaches on German texts exist to the best of our knowledge. 

Since we can only rely on very limited previous work, we will continue with a detailed discussion of the methods to encourage further research in this field.

\section{Method}
\subsection{Labor market data}\label{chap:data}

Labour markets are complex fields with diverse data structures and multiple applications, for example, connecting jobseekers to the right training or job \cite{felsenstein2006introduction}. For this reason the multilingual classification of European  Skills, Competences, Qualifications and Occupations (ESCO) is a good example for the central role of ontologies in this field, see \cite{de2015esco}. However, ESCO cannot provide all details of local labor market needs and does not provide links to other hierarchies of skills. For example, in German-speaking countries, other taxonomies of occupations and skills are widely used. Thus, when discussing data for occupational qualifications and certificates, we need to consider multiple data sets.

The International Standard Classification of Occupations (ISCO)\footnote{See \url{https://www.ilo.org/public/english/bureau/stat/isco/isco08/}.} 2008 was developed by the International Labour Organization (ILO) and was published in 1958, 1968, 1988 and as its recent version in 2008. It was also used within the European Union (EU), and some German-speaking countries (Germany, Austria, Switzerland) have developed a specific version of the ISCO 2008. ISCO also maps to the ontology ``European Skills, Competences, Qualifications and Occupations'' (ESCO) which also adds skills and competences to ISCO. 

In Germany, the ``Klassifikation der Berufe'' (KldB)\footnote{See \url{https://statistik.arbeitsagentur.de/DE/Navigation/Grundlagen/Klassifikationen/Klassifikation-der-Berufe/KldB2010-Fassung2020/KldB2010-Fassung2020-Nav.html}.} or DKZ is the reference for IAB (Institut für Arbeitsmarkt- und Berufsforschung) and the German Federal Employment Agency (Bundesagentur für Arbeit - BA). The most recent version is the 2020 revision of KldB 2010, which was completely redeveloped and makes previous versions from 1988 and 1992 deprecated. It was developed to be compatible to ISCO-08. 
This data are  part of the matching process in the BA and are integrated into other IT applications. However, while part ``B'' of KldB is dedicated to occupations, part ``C'' covers continuing professional development, ``K'' skills, and ``A'' higher education. All these parts are important to describe the access to education and training, see Table \ref{tab1} for details.

\begin{table}[t]
\caption{Data sets used to describe the access to education and training.}\label{tab1}
\centering
\begin{tabular}{|l|l|l|}
\hline
Data set & Content  & Entries\\
\hline
A &  Higher Education &  \multicolumn{1}{r|}{797}\\
B & Occupations &  \multicolumn{1}{r|}{33,802} \\
C & Continuing professional development &  \multicolumn{1}{r|}{542} \\
K & Skills &  \multicolumn{1}{r|}{9,078} \\
\hline
\end{tabular}
\end{table}

Part A mainly describes general academic education. This data is complex, since every federal state in Germany has a different approach, for example ``Berufliche Gymnasien'' do not exist in every state. In addition, some older terms are frequently used, for example ``Realschulabschluss'' instead of ``Sekundarabschluss I''. More precisely, 23 terms in A use ``Realschulabschluss'' as synonym. Thus, mapping these 797 entries to match the text description will be a first challenge.

Part B offers a wide description of 33,802 occupations. However, not all of them are relevant, since we focus on initial and continuing vocational education and training. Thus, we will add generic terms for describing the need for a previous training (B+) and professional experience (BE). For the same reason, part C is added for the sake of completeness. For example, very little trainings will \emph{only} focus on master craftsman or master craftswoman. 

We will now describe some existing taxonomies for skills in the German language and discuss why we limit our approach to BA data. Our first example is the European Classification for Skills, Competences, Qualifications and Occupations (ESCO). It provides a multi-language hierarchy of skills and competences (and in addition qualifications and occupations) containing a full text description, scope notes and comprising examples. 
Gonzalez et al. state, that only few works have described the analysis and use of ESCO, see \cite{gonzalez2020entity}.
Some work has been carried out for semantic interoperability between skills and labour market documents, which was initially promised by ESCO \cite{le2014esco}.  Other scholars tried to use data from ESCO and Wikidata for text mining on scientific literature, see \cite{gonzalez2020entity} or for curriculum analytics, see \cite{kitto2020towards}. Recent research has provided a generic mining and mapping approach \cite{fareri2021skillner} and automated ontology alignment for  ESCO and the English O*NET \cite{neutel2021towards}. 

But ESCO is not the only skill systematic available in German language, as we will discuss now. Krebs et al. provide a brief overview about the challenges when working with these resources \cite{krebs2022qube}. First, the competencies  must be relevant to work. 
Competencies and skills must be at a level of abstraction which can be used to distinguish between jobs. 
This means that competencies needs to be defined so broadly that they usually appear in more than one individual workplace. 
Competencies must offer added value compared to other existing classifications, which means they must be  multidisciplinary -- but at the same time, however, they should not be limited to a pure measurement of cognitive skills, as already implied in the qualification or requirement level. Obviously, again, this highlights the interdisciplinary challenges: A mapping to skills is not only a technical, semantic link between two resources, but should also keep the information and abstraction layers. 

While ESCO is funded by the EU and openly available, other resources are more restricted in their usage. For example, both employment agencies in Germany and Austria provide their own resources for structuring skills and linking them to occupations: the BERUFENET is developed by the German Federal Employment Agency (BA) and the `Klassifikation des Arbeitsmarktservice' 
is provided by the Austrian Labour Market Service (AMS). Both taxonomies are available via an online platform and are not interoperable, see \cite{krebs2022qube}.



Another concept of German skills was introduced by  MYSKILLS, see \cite{koppl2020fachkraftemangel,rittberger2022digitale}. It is primarily intended to define and discuss skills needed for a particular profession in the context of ``Unlocking the Potential of Migrants in Germany'' \cite{bergseng2019getting}. It contains a hierarchical definition of skills for ca. 30 professions as unstructured resources (data is available as PDF). The skills have an identifier and a description. 
Skills are also used in the field of adult training and lifelong learning. The model GRETA is special, because it also introduces several levels of skills, see \cite{lencer2016kompetenzmodell,lencer2016greta}.

As we can see, all different concepts have (a) different application purposes, (b) different data structures, (c) a different level of machine-readability and (d) are in general not interoperable. 
Several attempts were made for a standardization in this field. Konert et al. provided a  different perspective for the modelling of competences, see \cite{konert2019digitales}. They underlined validity and reliability as primary challenges for skill definitions. Rentzsch and Staneva provided a play for combining taxonomies and ontologies with data-driven methods in the field of skills matching and skills intelligence, see \cite{rentzsch2020skills}. 
Focusing on the modelling of skills, \cite{dahlmeyer2020semantic} introduced HyperCMS, an approach for semantic modelling in knowledge graphs.

Summarizing, for this version we decided to use the skills from BERUFENET, because they are included in the data used for BERUFENET database (see below) and form the most generic list. However, for future versions, including generic skills from ESCO and AMS will be key. 

\subsection{Data}\label{sec:data}

\begin{figure}[t] 
	\centering
	\includegraphics[width=0.45\textwidth]{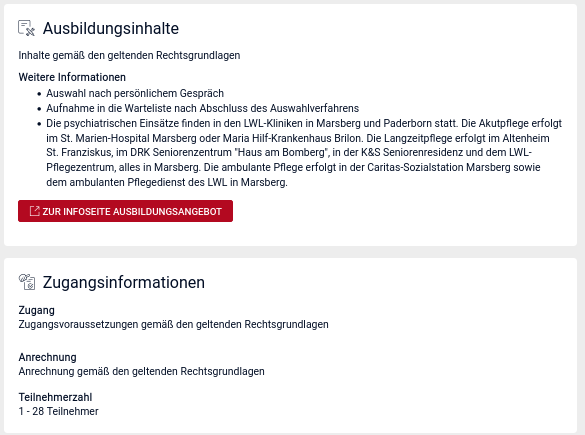}
	\caption{A screenshot of KURSNET data entries with two sections: training content (``Ausbildungsinhalt'') and access information (``Zugangsinformationen'') with the subsection access (``Zugang''). }
	\label{img:screen}
\end{figure}

Our first evaluation data was obtained from the data portal KURSNET of BA\footnote{See \url{https://www.arbeitsagentur.de/kursnet}.}. We retrieved data for initial and continuing vocational education and training, which leads to 23,460 (training),  and 66.549 (re-training) results in February 2023. Within the advertisement text, we only considered the section 	access information (``Zugangsinformationen'') with the subsection access (``Zugang'') which usually holds natural language describing the access limitations, see Figure \ref{img:screen}. 
\begin{table}[t]
\caption{Overview of data to evaluation the approach and their characteristic data content. ED refers to general school and education degrees and school-leaving certificates, PE to  professional experience, PA to previous apprenticeship and S to skills. Here, + indicates a wide usage of entities, - to an average usage and --- to nearly no usage.   }\label{tab0}
\centering
\begin{tabular}{|l|l|r|c|c|c|c|}
\hline
Data set & Source & Size                                                            &  ED & PE & PA & S\\
\hline
Re-trainings    & KURSNET                   & 120                                   & + & + & - & -\\
Training        & KURSNET           & 120                                           & + & + & --- & -\\
Continuing professional development & Weiterbildungsportal RLP & 120    & - & + & +  & +\\
\hline
\end{tabular}
\end{table} 

All data sets have a particular usage of terms from the previously described sets, see Table \ref{tab0}. These texts may refer to specific and unspecific access restrictiveness. For example:
\begin{quote}
Personal counseling interview, PROFILE aptitude test, at least secondary school leaving certificate required 
[Persönliches Beratungsgespräch, PROFIL-Eignungstest, mind. Hauptschulabschluss erforderlich]
\end{quote}
This text contains two unspecific access restrictions, a counselling interview, a non-standardized aptitude test and a lower secondary school-leaving certificate (``Hauptschulabschluss''). Other texts do not describe a specific access restriction at all:
\begin{quote}
good general education, manual dexterity, color aptitude, nationwide mobility  [gute Allgemeinbildung, handwerkliches Geschick, Farbtauglichkeit, bundesweite Mobilität]
\end{quote}
No standardized qualifications for general education, being good with hands, color blindness or mobility exist. This text cannot be mapped to the data described above. However, we selected 120 advertisements each to create a gold standard. This curation process was done manually. 

A second dataset to evaluate the proposed approach was obtained from ``Weiterbildungsportal Rheinland-Pfalz''\footnote{See \url{https://weiterbildungsportal.rlp.de/}.}. Here, the data mainly focuses on continuing professional development. See Table \ref{tab0} for a detailed overview on how this data set differs from the previously discussed. Thus, the access information is slightly different and often require one particular vocational education:
\begin{quote}
Successfully completed training in a recognized three-year commercial apprenticeship in the trade ... [Eine mit Erfolg abgeschlossene Ausbildung in einem anerkannten dreijährigen kaufmännischen Ausbildungsberuf im Handel ...]
\end{quote}
While this data is not in the primary focus of this work, which mainly focuses on trainings and re-trainings, it will show if our approach generalizes. 

Since the data is copyrighted, we can only publish a small set of the evaluation data which is available at \url{https://github.com/TM4VETR/DetectAccessToEducation}.

\subsection{Workflow}

\begin{figure}[t] 
	\centering
	\includegraphics[width=0.61\textwidth]{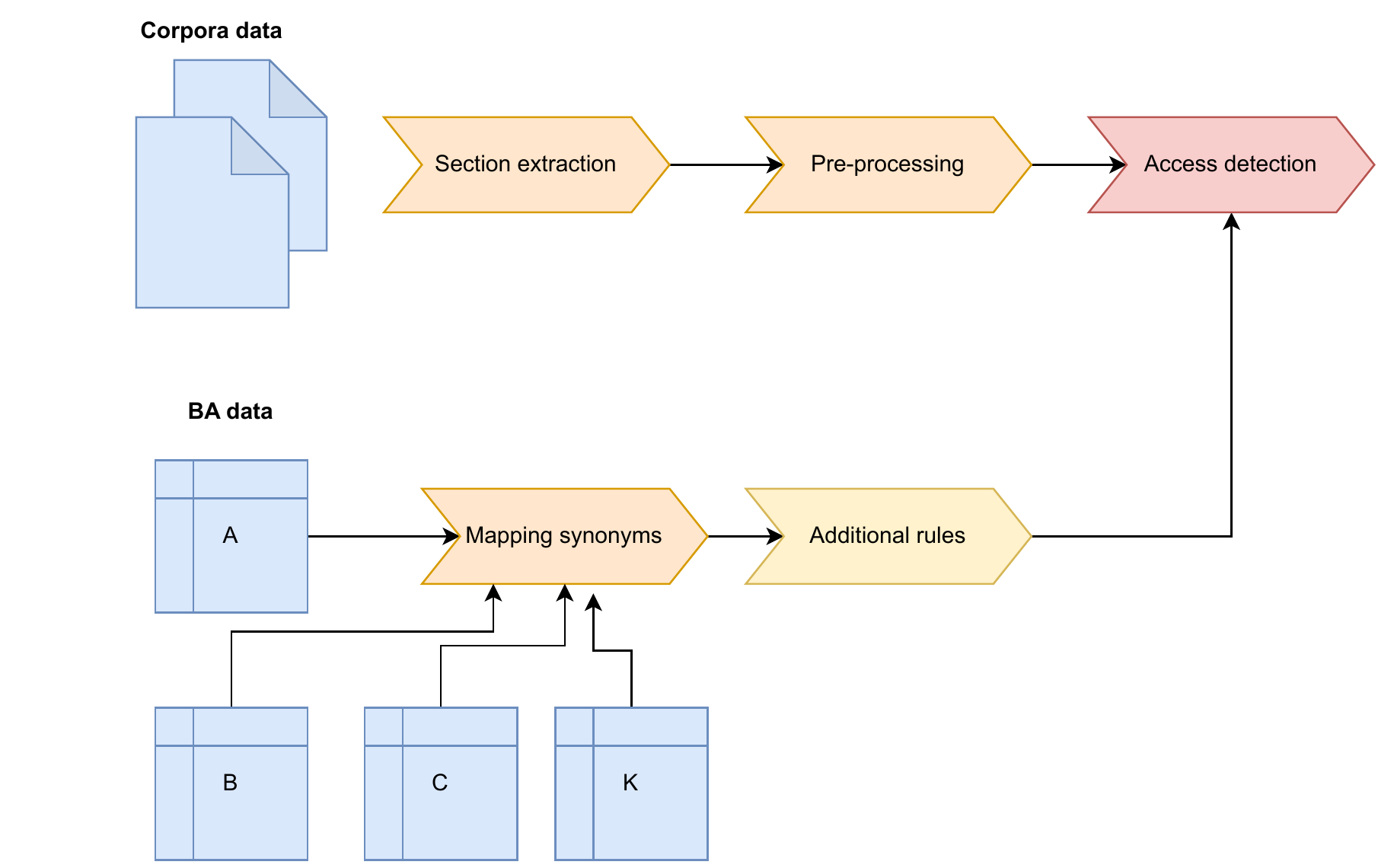}
	\caption{Flow-diagram of the processing workflow. }
	\label{img:pipe}
\end{figure}

The workflow comprises two independent parts, building the mappers and processing the corpora data. All parts have been developed using Python 3.9.2, spaCy 3.3.0, an advanced NLP library, and beautifulsoup4 4.9.3. Extracting the section from KURSNET website was done by extracting the content of the \texttt{<p>}-tag. However, we publish a generic method that can be used on any textual data. The pre-processing is done using spaCy using a tokenizer, sentence splitter, NER and the matcher. We use three different matchers to detect access information in the text. We use a \emph{PhraseMatcher} to match the large lists of entities described in sets B, C and K. For A, we build additional mappings and add synonyms. However, for additional rule-based matching, we use the \emph{Matcher} class. For more technical details, we refer to \cite{hernandez2018pyrata} or the spaCy documentation. We provide a full testing environment at \url{https://github.com/TM4VETR/DetectAccessToEducation}. 

We will now continue with a detailed discussion of how to map synonyms and qualifications in data sets A and B and how to build the additional rules.

\subsection{Mapping synonyms in education} 

We will now describe the algorithmic approach to map synonyms in data sets A and B containing  a list of general academic education and occupations, see Algorithm \ref{alg:step-1}. The data in A obtained from BA usually describes a general school and an education degree. 
The first part of each line contains the ID of a term (e.g. ``A 1.SAN-21.1''), the second the information. Thus, the first rule splits the degree (in this case ``allgemeine Hochschulreife'') and the school. See line \ref{line:klammer} in Algorithm \ref{alg:step-1}. However, some terms only list courses which refer to a training preparation (``Vorbereitung'') which need to be omitted since they do not provide the formal qualification. 
In addition, we provide a rich list of synonyms, for example ``mittlerer Schulabschluss'' to ``Realschulabschluss'', ``Fachoberschulreife'', ``Sekundarabschluss I'' and ``mittlere Reife''. Some terms are deprecated but are still widely used. In line \ref{line:syn} in Algorithm \ref{alg:step-1}, we additional synonyms are added.  

The situation is more complex when focusing on data set B. Here, male and female titles are provided together, and optional regular information be provided. For example:
\begin{quote}
B 27302-902|Produktionstechnologe/Produktionstechnologin|\\
B 28222-905|Fachpraktiker/Fachpraktikerin für Näherei und Schneiderei (§66 BBiG/§42r HwO)|
\end{quote}
In the first case, a complete male and female title is split by a slash. In the second case, only the first part of the title is split. In line \ref{line:fem} the two functions \emph{male} and \emph{female} are used to describe the extraction process.

\begin{algorithm}[t]
\begin{algorithmic}[2]
\caption{\textsc{Mapping Synonyms}} \label{alg:step-1}
\REQUIRE List $L$ of entries from data sets A and B 
\REQUIRE List $S$ of synonyms for general school and education degrees
\ENSURE List $L'$ with synonyms 
\STATE $L'=\emptyset$
\FOR {every $l \in L$ } 
\FOR{$s\in S$} \label{line:syn}
\IF{$s\in l$}
\STATE $L' \leftarrow $ [$id(l),s$]
\ENDIF
\ENDFOR
\IF{"("$\in l$}\label{line:klammer}
\STATE $L' \leftarrow $ [$id(l),split(l,"(",0)$]
\STATE $L' \leftarrow $ [$id(l),split(l,"(",1).remove(")")$]
\ENDIF
\IF{"/"$\in l$}\label{line:fem}
\STATE $L' \leftarrow $ [$id(l),male(l)$]
\STATE $L' \leftarrow $ [$id(l),female(l)$]
\ENDIF
\ENDFOR
\RETURN $L\cup L'$
\end{algorithmic}
\end{algorithm}


\subsection{Rule-based matching}

The complexity of natural text leads to several challenges. We find multiple ways to describe professional experience (Berufserfahrung), school-leaving certificates (Schulabschluss) or technical and vocational education and training. 

\begin{figure}[t] 
	\centering
	\includegraphics[width=0.61\textwidth]{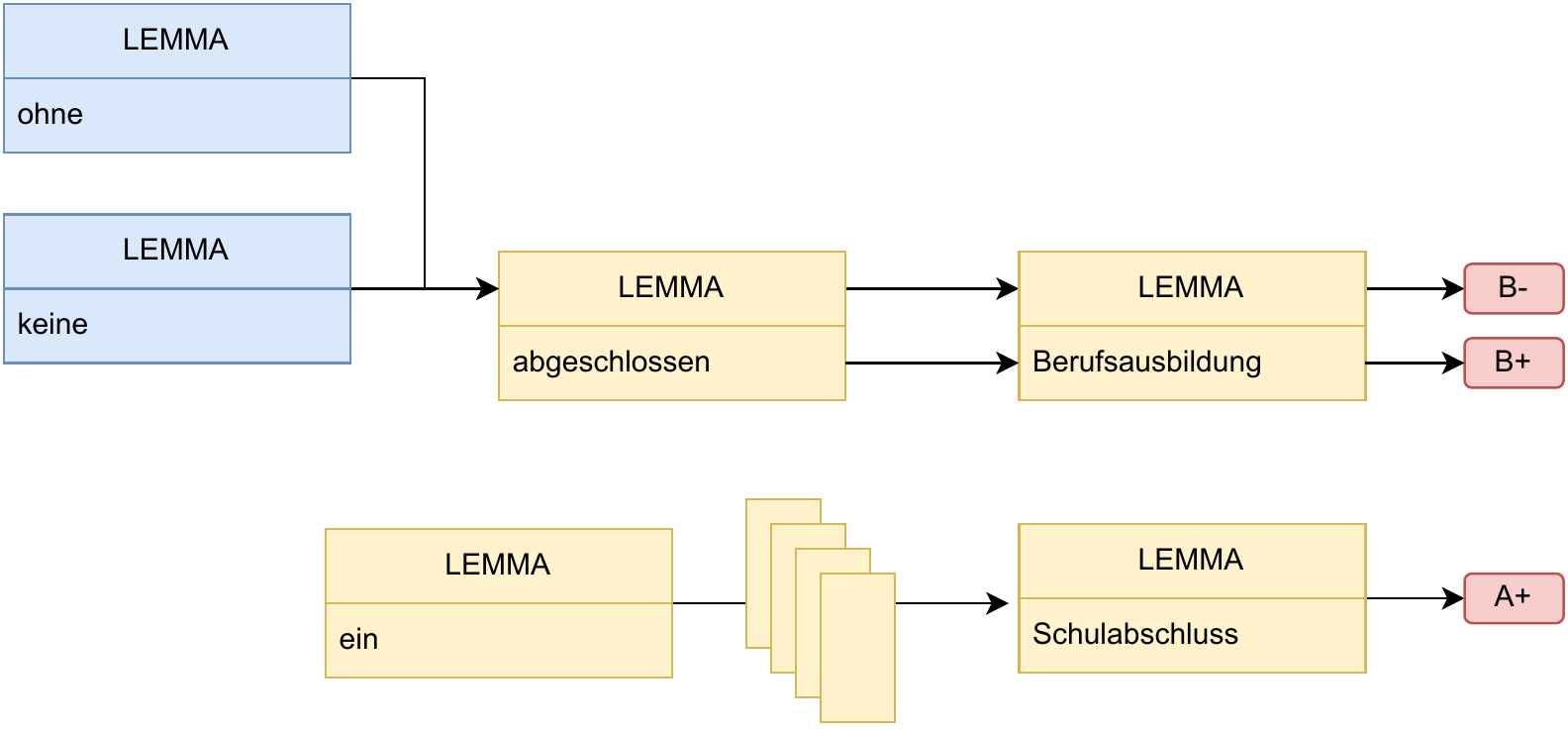}
	\caption{Two example rules to detect school-leaving certificates (Schulabschluss, bottom) or technical and vocational education (top). All rules have positive (e.g. A+) and negative (e.g. A-) results. An existing negative result will remove a positive one.}
	\label{img:rule}
\end{figure}

Professional experience is mentioned in several more or less complex sentences:
\begin{quote}
But also dropouts or persons \textbf{with work experience} in other fields should inform themselves about the retraining offer if they are interested...\\
Completed vocational training or \textbf{work experience}...\\
\textbf{work experience} desirable (1 year), but not a must....
\end{quote}
In the last sentence, experience is only preferable (``wünschenswert''), but not a condition. School-leaving certificates are usually found in a similar context and setting. However, as described above, since particular certificates are usually named explicitly the extraction itself is not that challenging. See Figure \ref{img:rule} (bottom) for an illustration of a rule matching several complex phrases. However, extracting previous vocational education is more challenging. Some texts directly mention the prerequisite:
\begin{quote}
Minimum requirement: \textbf{vocational training}
\end{quote}
Usually, a vocational education is embedded in an or-clause stating that it is not the only way to access the training:
\begin{quote}
If possible, completed \textbf{vocational training} or ... \\
At least 4 years of professional activity or completed \textbf{vocational training}....
\end{quote}
However, a sentence matched by a rule for negative phrases (... \textbf{keine} abgeschlossene Berufsausbildung...) might also be matched by a positive rule (... eine abgeschlossene Berufsausbildung...). Thus, we define positive and negative results (B+, B-) and an existing negative result will remove a positive one, see Figure \ref{img:rule} (top) for an illustration.

We provide several rules for frequently used structures like ``with our without'' (... ohne oder mit...) or negative clauses. Our rule-based approach does not differentiate between or- and and-clauses. However, we observed that most descriptions use or-clauses. Thus, in these cases, all detected qualifications give access to the training offered. Another limitation is the naming of language qualifications (``angemessene Deutschkenntnisse'' or ``Deutschkenntnisse ... B2''). The biggest challenge is that these qualifications cannot be exactly matched in KldB which only offers ``Deutsch - verschiedene Niveaustufen'' (A 8.11) or Grundstufe, Mittelstufe and Oberstufe. Second, we only found requirements for German, not for any other language. Thus, as a preliminary solution, we map these rules to ``A 8.11''.

\subsection{Limitations}

As described above, we heavily rely on the data from BA. While  school-leaving certificates and occupations are regulated by the German government, skills and other tests are usually not well-defined and the BA-data omits several of them. We decided to leave the missing data for this version, as it will require more analysis of their nature. For example, we found ``ZAUG-Eignungsfeststellung'', or ``CDT-Eignungstest für IT-Berufe''. Thus, unspecific access restrictions, e.g. a counselling interview, and non-standardized tests are currently not considered. 

\section{Experimental results}

As described in Section \ref{sec:data} we rely on data from BERUFENET database for initial and continuing vocational education and training. Our gold standard to evaluate the output of our approach are 120 sets  for training (Berufsausbildung) and re-training (Umschulung) advertisements. A third data set focusing on continuing professional development was obtained from ``Weiterbildungsportal Rheinland-Pfalz''.   To analyze the quality of this network, we can calculate precision and recall:
\[ Precision = \frac{TP}{TP+FP},\;\; Recall = \frac{TP}{TP+FN}\]
Here, $TP$ refers to true positive results, $FP$ to all false positive results, and $FN$ are false negative results. With precision and recall, we can compute the $F_1$-score, which is as a weighted average of the precision and the recall:
\[F_1 = 2 \cdot \frac{\textit{Precision} \cdot \textit{Recall}}{\textit{Precision} + \textit{Recall}}\]

\begin{table}[t]
\caption{Overview of metrics and results: Precision, recall and $F_1$ scores on different data sets.}\label{tab2}
\centering
\begin{tabular}{|l|r|r|r|}
\hline
Data set & Precision & Recall & $F_1$-score \\
\hline
Re-trainings    & 0.86 & 0.95 & 0.91 \\
Training        & 0.99 & 0.95 & 0.97 \\
Continuing professional development & 0.47 & 0.30 & 0.36\\
\hline
\end{tabular}
\end{table}

For a detailed overview, we refer to Table \ref{tab2}. It is apparent from this table that for the first two data sets, the $F_1$-score is high. What is interesting in this data is that the precision on re-training data is much lower than for trainings. This is mainly due to the wrong detection of occupations. For example, one advertisement asks interested people to call the consultant (Fachberater) in case of queries:
\begin{quote}
... and have their personal access requirements checked by our \textbf{consultants}. [... und ihre persönlichen Zugangsvoraussetzungen durch unsere \textbf{Fachberater/innen} prüfen lassen.]
\end{quote}
However, our approach detects several occupations to access this training: ``Fachberater/ Fachberaterin für Softwaretechniken/ Finanzdienstleistungen/  Vertrieb/ integrierte Systeme'' and even a different position in commerce: ``Fachverkäufer/ Fachberater / Fachverkäuferin/ Fachberaterin - Bau-/Heimwerkerbedarf''. Another difficulty are legal aspects. For example, access to a training might be easier with support (so-called ``Bildungsgutschein''), thus they are mentioned in the advertisement: ``SGB II/SGB III (Bildungsgutschein)''. In this case, ``SGB III'' also refers to the  unemployment insurance and the skill 030400-011 (Sozialgesetzbuch III) is detected. However, the recall is in general sufficient high. 

Interestingly, the results change on data set three. When comparing the results to the other sets, neither precision nor recall show a working approach. In this data, more frequently previous trainings are required. First, we will discuss a complex example:
\begin{quote}
Successful completion of training in a recognized three-year commercial trade occupation or at least one year's professional experience, or successful completion of the final examination for sales assistant or in another recognized trade occupation, followed by at least five years' professional experience if the above requirements are not met. 
\end{quote}
This text describes the following requirements: An apprenticeship in trade, \emph{or} one year professional experience, \emph{or} apprenticeship as shop assistant, \emph{or} any other apprenticeship \emph{and} five years professional experience. However, while only few apprenticeships as shop assistant (Verkäufer/Verkäuferin) exist, it is unclear, which apprenticeships are in trade (either ``kaufmännisch'' or ``im Handel''). In any case, our approach fails to precisely enumerate them. In general, we found no reference to one particular training, but only general statements:
\begin{quote}
A \textbf{completed electrotechnical training} oriented to the specifications of TRBS 1203....\\
A completed \textbf{technical vocational training}... \\
\end{quote}
This shows that our approach is able to detect some elements to access education and training in German language: Mapping synonyms in education works well and extracts general school and education degrees needed for a particular training. The rule-based matching approach to detect professional experience (Berufserfahrung), a previous apprenticeship and school-leaving certificates (Schulabschluss) as well as extracting skills from the BA-list shows promising results. However, extracting particular occupations and trainings does not work as expected. Our approach either extracts occupations which are not a prerequisite or cannot dissolve groups of trainings. 

Summarizing, the proposed method works well for data which does not use occupational or training references, namely training and pre-training advertisements, but does not yet generalize.

\section{Conclusion and Outlook}

\begin{figure}[t] 
	\centering
	\includegraphics[width=0.35\textwidth]{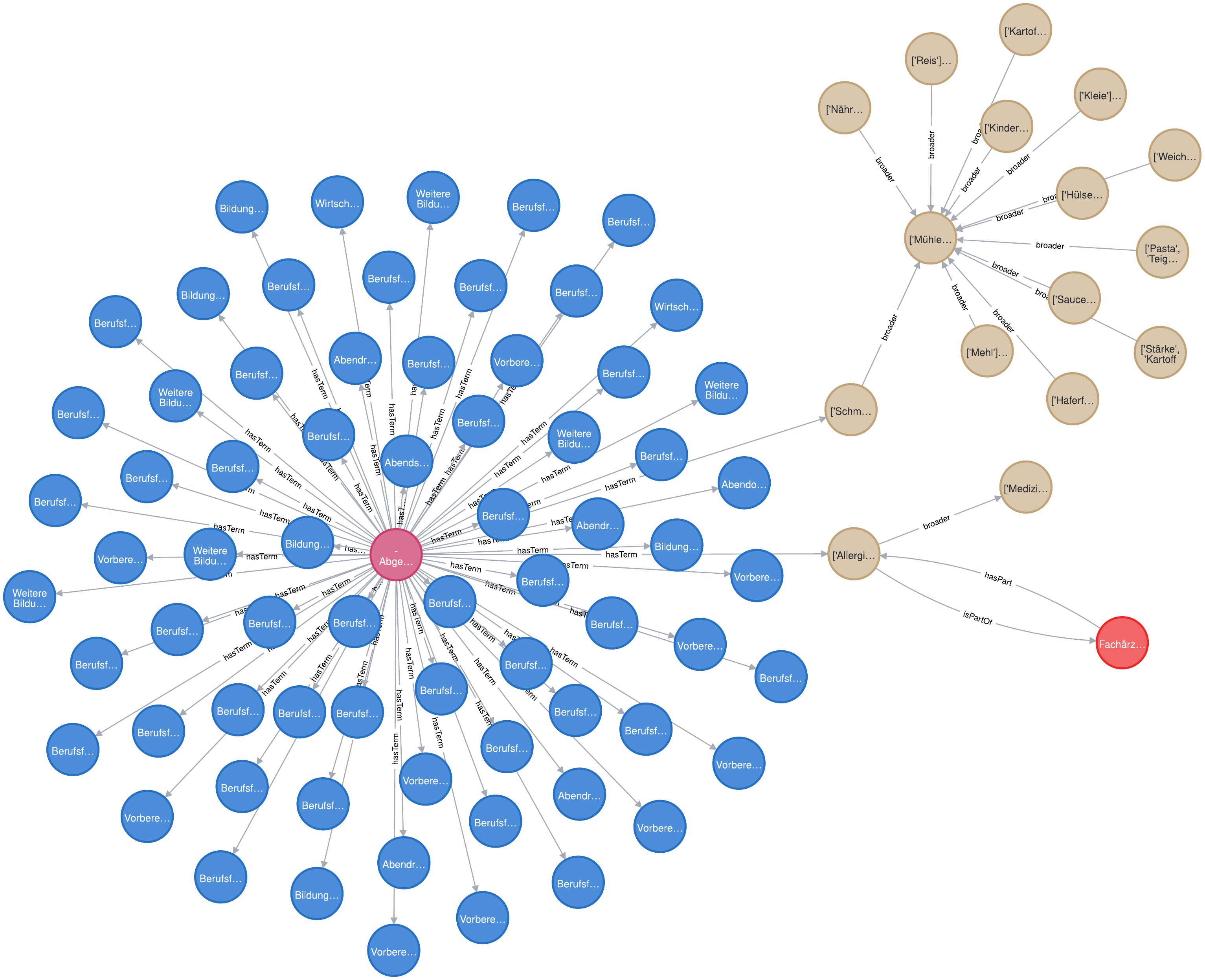} \quad\quad
        \includegraphics[width=0.35\textwidth]{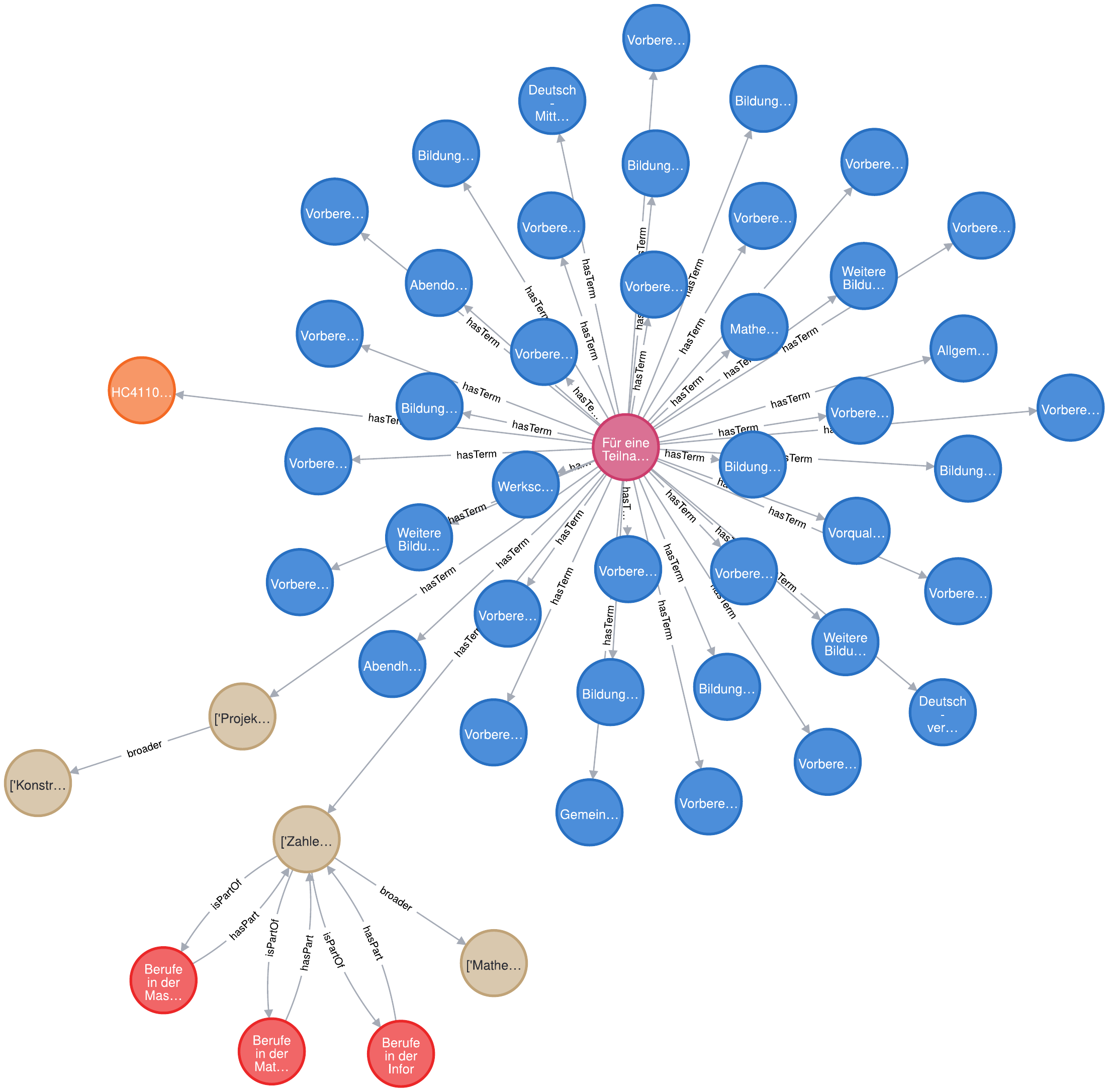} 
	\caption{Two example outputs of the proposed workflow embedded in the knowledge graph on labour market data. The trainings advertisement (purple) is in the center, blue nodes refer to school qualifications, brown nodes to skills, orange to vocational education. The skills can be linked to occupations (red). }
	\label{img:graph}
\end{figure}

Vocational education and training, re-training and advanced vocational qualification are key to meet the requirements of the labour market of tomorrow. To match training seekers and offers, we presented a novel approach towards the automated detection of access to education and training in German training offers and advertisements.  

We focused on (a) general school and education degrees and school-leaving certificates, (b) professional experience (``Berufserfahrung''), (c) a previous apprenticeship and (d) skills. Our novel approach combined several methods: First, we provided a mapping of synonyms in education combining different qualifications and adding deprecated terms. As our results and discussion show, this works well for the evaluation data. 

Second, we provided a rule-based matching to identify the need for professional experience or apprenticeship. However, as discussed, not all access requirements can be matched due to incompatible data schemata or non-standardizes requirements, e.g. initial tests or interviews. However, extracting particular occupations and trainings does not work as expected. Our approach either extracts occupations which are not a prerequisite or cannot dissolve groups of trainings. 

The presented approach shows that on the one hand the initial data to map and on the other hand the text structures have a great impact on the efficiency and precision of the presented approach. The experimental results show promising results for two data sets (trainings and re-trainings), but further research is necessary, in particular on the identified shortcomings, to build a generic tool suitable for different input data. 

However, we have integrated the presented approach in a generic workflow to analyse labour market data in a knowledge graph, see Figure \ref{img:graph}. We propose that further data integration (e.g. of other skill taxonomies) is needed towards modelling and analysis of training advertisements. This approach towards the detection of access to education and training in Germany is only a first step.


%
%
%
 \bibliographystyle{splncs04}
%
\bibliography{lit}

\begin{thebibliography}{10}
\providecommand{\url}[1]{\texttt{#1}}
\providecommand{\urlprefix}{URL }
\providecommand{\doi}[1]{https://doi.org/#1}

\bibitem{ben2011recruiting}
Ben~Abdessalem, W.K., Amdouni, S.: E-recruiting support system based on text
  mining methods. International Journal of Knowledge and Learning
  \textbf{7}(3-4),  220--232 (2011)

\bibitem{berkesewicz2021inferring}
Ber{\k{e}}sewicz, M., Pater, R.: Inferring job vacancies from online job
  advertisements. Publications Office of the European Union (2021)

\bibitem{bergseng2019getting}
Bergseng, B., Degler, E., L{\"u}thi, S.: Getting migrants ready for vocational
  education and training in germany. Unlocking the Potential of Migrants in
  Germany  (2019)

\bibitem{dahlmeyer2020semantic}
Dahlmeyer, M.P.: Semantic competence modelling--observations from a hands-on
  study with hypercmp knowledge graphs and implications for modelling
  strategies and semantic editors. In: Proceedings of DELFI Workshops 2020.
  Gesellschaft f{\"u}r Informatik eVz (2020)

\bibitem{de2015esco}
De~Smedt, J., le~Vrang, M., Papantoniou, A.: Esco: Towards a semantic web for
  the european labor market. In: Ldow@ www (2015)

\bibitem{degenhardt2018kompetenzen}
Degenhardt, S.: Kompetenzen f{\"u}r eine digitalisierte
  arbeitswelt--anforderungen an aus-und weiterbildung. In: Digitaler Wandel in
  der Sozialwirtschaft. pp. 259--272. Nomos Verlagsgesellschaft mbH \& Co. KG
  (2018)

\bibitem{dobischat2020organisation}
Dobischat, R., D{\"u}sseldorff, K.: {Organisation, Recht und Finanzierung der
  beruflichen Weiterbildung}. Handbuch Berufsbildung pp. 579--595 (2020)

\bibitem{dobischat2019digitalisierung}
Dobischat, R., K{\"a}pplinger, B., Molzberger, G., M{\"u}nk, D.:
  Digitalisierung und die folgen: Hype oder revolution? Bildung 2.1 f{\"u}r
  Arbeit 4.0? pp. 9--24 (2019)

\bibitem{dutt2017systematic}
Dutt, A., Ismail, M.A., Herawan, T.: A systematic review on educational data
  mining. Ieee Access  \textbf{5},  15991--16005 (2017)

\bibitem{fareri2021skillner}
Fareri, S., Melluso, N., Chiarello, F., Fantoni, G.: Skillner: Mining and
  mapping soft skills from any text. Expert Systems with Applications
  \textbf{184},  115544 (2021)

\bibitem{felsenstein2006introduction}
Felsenstein, D., McQuaid, R.W.: Introduction to the special issue: linking
  demand and supply in local labor market research. The Annals of Regional
  Science  \textbf{40},  389--392 (2006)

\bibitem{gonzalez2020entity}
Gonz{\'a}lez, L., Garc{\'\i}a-Barriocanal, E., Sicilia, M.A.: Entity linking as
  a population mechanism for skill ontologies: Evaluating the use of esco and
  wikidata. In: Research Conference on Metadata and Semantics Research. pp.
  116--122. Springer (2020)

\bibitem{graf2021advanced}
Graf, L., Lohse, A.P.: Advanced skill formation between vocationalization and
  academization: The governance of professional schools and dual study
  programmes in germany. Governance Revisited. Challenges and Opportunities for
  Vocational Education and Training; Gonon, P., B{\"u}rgi, R., Eds  (2021)

\bibitem{helmrich2016digitalisierung}
Helmrich, R., Tiemann, M., Troltsch, K., Lukowski, F., Neuber-Pohl, C.,
  Lewalder, A.C., Gunturk-Kuhl, B.: Digitalisierung der Arbeitslandschaften:
  keine Polarisierung der Arbeitswelt, aber beschleunigter Strukturwandel und
  Arbeitsplatzwechsel. No.~180, Wissenschaftliche Diskussionspapiere (2016)

\bibitem{hermes2016stellenanzeigenanalyse}
Hermes, J., Schandock, M.: Stellenanzeigenanalyse in der
  qualifikationsentwicklungsforschung. Die Nutzung maschineller Lernverfahren
  zur Klassifikation von Textabschnitten. Bundesinstitut f{\"u}r Berufsbildung,
  Bonn  (2016)

\bibitem{hernandez2018pyrata}
Hernandez, N., Hazem, A.: Pyrata, python rule-based feature structure analysis.
  In: Proceedings of the Eleventh International Conference on Language
  Resources and Evaluation (LREC 2018) (2018)

\bibitem{kitto2020towards}
Kitto, K., Sarathy, N., Gromov, A., Liu, M., Musial, K., Buckingham~Shum, S.:
  Towards skills-based curriculum analytics: Can we automate the recognition of
  prior learning? In: Proceedings of the tenth international conference on
  learning analytics \& knowledge. pp. 171--180 (2020)

\bibitem{konert2019digitales}
Konert, J., Buchem, I., Stoye, J.: Digitales kompetenzverzeichnis mit
  technologien des semantic web. In: Proceedings of DELFI Workshops 2019.
  Gesellschaft f{\"u}r Informatik eVz (2019)

\bibitem{kopparapu2010automatic}
Kopparapu, S.K.: Automatic extraction of usable information from unstructured
  resumes to aid search. In: 2010 IEEE International Conference on Progress in
  Informatics and Computing. vol.~1, pp. 99--103. IEEE (2010)

\bibitem{koppl2020fachkraftemangel}
K{\"o}ppl, C.: Fachkr{\"a}ftemangel durch quereinsteiger abfedern.
  Wirtschaftsinformatik \& Management  \textbf{12}(5),  382--385 (2020)

\bibitem{kovalev2020educational}
Kovalev, S., Kolodenkova, A., Muntyan, E.: Educational data mining: current
  problems and solutions. In: 2020 V International Conference on Information
  Technologies in Engineering Education (Inforino). pp.~1--5. IEEE (2020)

\bibitem{krebs2022qube}
Krebs, B., Maier, T.: {Die QuBe-Kompetenzklassifikation als verdichtende
  Perspektive auf berufliche Anforderungen}. Tech. rep., Wissenschaftliche
  Diskussionspapiere (2022)

\bibitem{kreuzer2018visualisierung}
Kreuzer, C.: Visualisierung der opportunity recognition-kompetenz von
  industriekaufleuten. Zeitschrift f{\"u}r Berufs-und Wirtschaftsp{\"a}dagogik
  \textbf{114}(2),  247--271 (2018)

\bibitem{germany2019national}
of~Labour, G.F.M., of~Education, S.A.B.G.F.M., (BMBF), R.: {National Skills
  Strategy: continuing education and training as a response to digital
  transformation}  (2019)

\bibitem{lencer2016greta}
Lencer, S., Strauch, A.: Das greta-kompetenzmodell f{\"u}r lehrende in der
  erwachsenen-und weiterbildung. Abgerufen von https://www. die-bonn.
  de/doks/2016-erwachsenenbildung-02. pdf  (2016)

\bibitem{lencer2016kompetenzmodell}
Lencer, S., Strauch, A.: Ein kompetenzmodell f{\"u}r lehrende in der
  erwachsenen-und weiterbildung: erste ergebnisse aus dem projekt greta. DIE
  Zeitschrift f{\"u}r Erwachsenenbildung (4),  40--41 (2016)

\bibitem{mohamad2013educational}
Mohamad, S.K., Tasir, Z.: Educational data mining: A review. Procedia-Social
  and Behavioral Sciences  \textbf{97},  320--324 (2013)

\bibitem{neutel2021towards}
Neutel, S., de~Boer, M.H.: Towards automatic ontology alignment using bert. In:
  AAAI Spring Symposium: Combining Machine Learning with Knowledge Engineering
  (2021)

\bibitem{rentzsch2020skills}
Rentzsch, R., Staneva, M.: Skills-matching und skills intelligence durch
  kuratierte und datengetriebene ontologien. In: Proceedings of DELFI Workshops
  2020. Gesellschaft f{\"u}r Informatik eVz (2020)

\bibitem{rittberger2022digitale}
Rittberger, M.: Digitale bildung: Rolle und chancen einer
  forschungsinfrastruktureinrichtung. In: ZPID-Kolloquium 2019, Trier, Germany.
  ZPID (Leibniz Institute for Psychology Information) (2022)

\bibitem{romero2007educational}
Romero, C., Ventura, S.: Educational data mining: A survey from 1995 to 2005.
  Expert systems with applications  \textbf{33}(1),  135--146 (2007)

\bibitem{schiersmann2022weiterbildungsberatung}
Schiersmann, C.: Weiterbildungsberatung im kontext der nationalen
  weiterbildungsstrategie: Finanzielle und strukturelle aspekte. Hessische
  Bl{\"a}tter f{\"u}r Volksbildung  \textbf{72}(1),  43--53 (2022)

\bibitem{steeg2022wasserstoffwirtschaft}
Steeg, S.: {Die Wasserstoffwirtschaft in Deutschland: Folgen f\"ur Arbeitsmarkt
  und Bildungssystem; eine erste Bestandsaufnahme}  (2022)

\bibitem{le2014esco}
le~Vrang, M., Papantoniou, A., Pauwels, E., Fannes, P., Vandensteen, D.,
  De~Smedt, J.: Esco: Boosting job matching in europe with semantic
  interoperability. Computer  \textbf{47}(10),  57--64 (2014)

\bibitem{ziegler2012verwendung}
Ziegler, P.: Zur verwendung von berufsinformation im hinblick auf matching in
  deutschland und {\"o}sterreich. Tech. rep., AMS info (2012)

\bibitem{zimmermann2013youth}
Zimmermann, K.F., Biavaschi, C., Eichhorst, W., Giulietti, C., Kendzia, M.J.,
  Muravyev, A., Pieters, J., Rodr{\'\i}guez-Planas, N., Schmidl, R., et~al.:
  Youth unemployment and vocational training. Foundations and
  Trends{\textregistered} in Microeconomics  \textbf{9}(1--2),  1--157 (2013)

\end{thebibliography}
\end{document}